# Cat Swarm Optimization Algorithm - A Survey and Performance Evaluation

Aram M. Ahmed,[1,3] Tarik A. Rashid,[2] Soran Ab. M. Saeed[3]

[1]International Academic Office, Kurdistan Institution for Strategic Studies and Scientific Research, Sulaymaniyah 46001, Iraq.
[2]Computer Science and Engineering, University of Kurdistan Hewler, Erbil 44001, Iraq.
[3]Information Technology, Sulaimani Polytechnic University, Sulaymaniyah 46001, Iraq.

Correspondence should be addressed to Aram M. Ahmed; aramahmed@kissr.edu.krd

## Abstract

This paper presents an in-depth survey and performance evaluation of the Cat Swarm Optimization (CSO) Algorithm. CSO is a robust and powerful metaheuristic swarm-based optimization approach that has received very positive feedback since its emergence. It has been tackling many optimization problems and many variants of it have been introduced. However, the literature lacks a detailed survey or a performance evaluation in this regard. Therefore, this paper is an attempt to review all these works, including its developments and applications, and group them accordingly. In addition, CSO is tested on 23 classical benchmark functions and 10 modern benchmark functions (CEC 2019). The results are then compared against three novel and powerful optimization algorithms, namely Dragonfly algorithm (DA), Butterfly optimization algorithm (BOA) and Fitness Dependent Optimizer (FDO). These algorithms are then ranked according to Friedman test and the results show that CSO ranks first on the whole. Finally, statistical approaches are employed to further confirm the outperformance of CSO algorithm.

## 1. Introduction

Optimization is the process by which the optimal solution is selected for a given problem among many alternative solutions. One key issue of this process is the immensity of the search space for many real-life problems, in which it is not feasible for all solutions to be checked in a reasonable time. Nature-inspired algorithms are stochastic methods, which are designed to tackle these types of optimization problems. They usually integrate some deterministic and randomness techniques together, and then iteratively compare a number of solutions until a satisfactory one is found. These algorithms can be categorized into trajectory-based and population-based classes [1]. In trajectory-based types, such as a simulated annealing algorithm [2], only one agent is searching in the search space to find the optimal solution. Whereas, in the population-based algorithms, also known as Swarm Intelligence, such as Particle Swarm Optimization (PSO) [3], multiple agents are searching and communicating with each other in a decentralized manner to find the optimal solution. Agents usually move in two phases, namely Exploration and Exploitation. In the first one, they move on a global scale to find promising areas. While in the second one, they search locally to discover better solutions in those promising areas found so far. Having a trade-off between these two phases, in any algorithm, is very crucial because biasing towards either exploration or exploitation would degrade the overall performance and produce undesirable results [1]. Therefore, more than hundreds of swarm intelligence algorithms have been proposed by researchers to achieve this balance and provide better solutions for the existing optimization problems.

Cat Swarm Optimization (CSO) is a Swarm Intelligence algorithm, which is originally invented by Chu et al. in 2006 [4,5]. It is inspired by the natural behavior of cats and it has a novel technique in modeling exploration and exploitation phases. It has been successfully applied in various optimization fields of science and engineering. However, the literature lacks a recent and detailed review of this algorithm. In addition, since 2006 CSO has not been compared against novel algorithms i.e. it has been mostly compared with PSO algorithm while many new algorithms have been introduced since then. So, a question, which arises, is whether CSO competes with the novel algorithms or not? Therefore, experimenting with CSO on a wider range of test functions and comparing it with new and robust algorithms will further reveal the potential of the algorithm. As a result, the aims of this paper are: firstly, provide a comprehensive and detailed review of the state of art of CSO algorithm (see Figure 1), which shows the general framework for conducting the survey; secondly, evaluate the performance of CSO algorithm against modern metaheuristic algorithms. These should hugely help researchers to further work in the domain in terms of developments and applications.

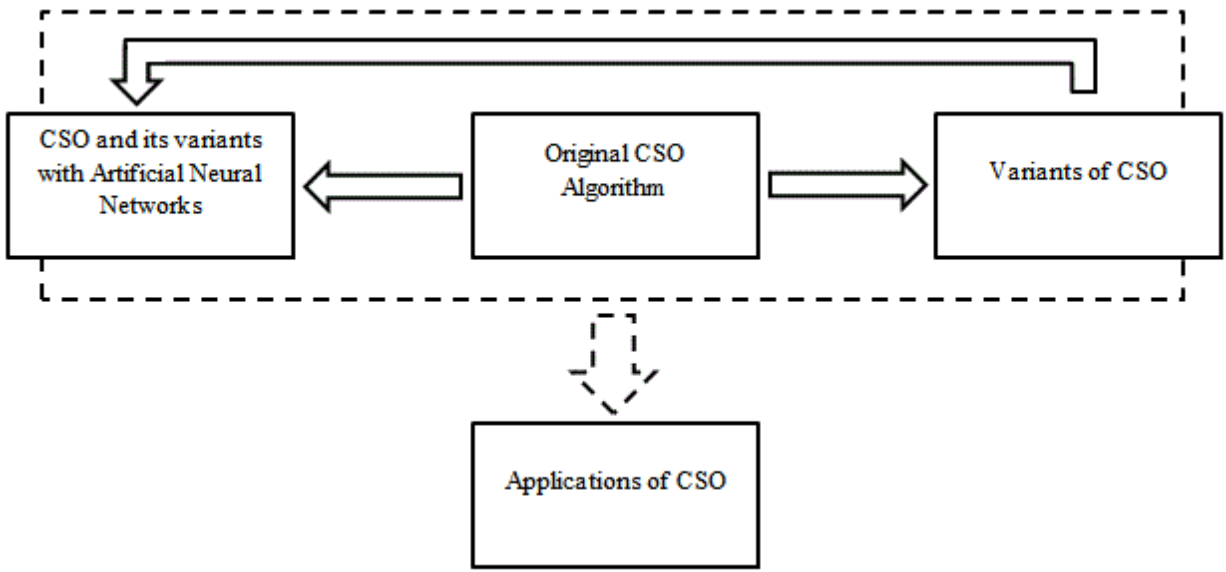

Figure 1: General framework for conducting the survey.

The rest of the paper is organized as follows; Section 2 presents the original algorithm and its mathematical modeling. Section 3 is dedicated to reviewing all modified versions and variants of CSO. Section 4 summarizes the hybridizing CSO algorithm with ANN and other Non-Metaheuristic methods. Section 5 presents applications of the algorithm and groups them according to their disciplinary. Section 6 provides performance evaluation, where CSO is compared against the Dragonfly algorithm (DA) [6], Butterfly optimization algorithm (BOA) [7] and Fitness Dependent Optimizer (FDO) [8]. Finally, section 7 provides the conclusion and future directions.

## 2. Original Cat Swarm Optimization Algorithm

The original Cat Swarm Optimization is a continuous and single-objective algorithm [4,5]. It is inspired by resting and tracing behaviours of cats. Cats seem to be lazy and spend most of their time resting. However, during their rests, their consciousness is very high and they are very aware of what is happening around them. So, they are constantly observing the surroundings intelligently and deliberately and when they see a target, they start moving towards it quickly. Therefore, the CSO algorithm is modeled based on combining these two main deportments of cats.

CSO algorithm is composed of two modes, namely tracing and seeking modes. Each cat represents a solution set, which has its own position, a fitness value and a flag. The position is made up of M dimensions in the search space and each dimension has its own velocity; the fitness value depicts how well the solution set (cat) is; and finally, the flag is to classify the cats into either seeking or tracing mode. Thus, we should first specify how many cats should be engaged in the iteration and run them through the algorithm. The best cat in each iteration is saved into memory and the one at the final iteration will represent the final solution.

2.1 The general structure of the algorithms: The algorithm takes the following steps in order to search for optimal solutions.

1. Specify the upper and lower bounds for the solution sets.
2. Randomly generate $N$ cats (solution sets) and spread them in the $M$ dimensional space in which each cat has a random velocity value not larger than a predefined maximum velocity value.
3. Randomly classify the cats into seeking and tracing modes according to $MR$. $MR$ is a mixture ratio, which is chosen in the interval of [0, 1]. So, for example, if a number of cats $N$ is equal to 10 and $MR$ is set to 0.2 then 8 cats will be randomly chosen to go through seeking mode and the other 2 cats will go through tracing mode.
4. Evaluate the fitness value of all the cats according to the domain-specified fitness function. Next, the best cat is chosen and saved into memory.
5. The cats then move to either seeking or tracing mode.
6. After the cats are going through seeking or tracing mode, for the next iteration, randomly redistribute the cats into seeking or tracing modes based on $MR$.
7. Check the termination condition, if satisfied; terminate the program, otherwise, repeat Step 4 to Step 6.

2.2 Seeking mode: This mode imitates the resting behavior of cats, where four fundamental parameters are playing important roles: seeking memory pool ($SMP$), seeking a range of the selected dimension ($SRD$), counts of dimension to change ($CDC$), and self-position considering ($SPC$). These values are all tuned and defined by the user through a trial-and-error method.

$SMP$ specifies the size of seeking memory for cats i.e. it defines number of candidate positions in which one of them is going to be chosen by the cat to go to, for example, if $SMP$ was set to 5 then for each and every cat 5 new random positions will be generated and one of them will be selected to be the next position of the cat. How to randomize the new positions will depend on the other two parameters that are $CDC$ and $SRD$. $CDC$ defines how many dimensions to be modified which is in the interval of [0, 1]. For example, if the search space has 5 dimensions and *CDC* is set to 0.2 then for each cat four random dimensions out of the five need to be modified and the other one stays the same. *SRD* is the mutative ratio for the selected dimensions i.e. it defines the amount of mutation and modifications for those dimensions that were selected by the *CDC*. Finally, *SPC* is a Boolean value, which specifies whether the current position of a cat will be selected as a candidate position for the next iteration, or not. So, for example, if the *SPC* flag is set to true then, for each cat, we need to generate (*SMP*-1) number of candidates instead of *SMP* numbers as the current position is considered as one of them. Seeking mode steps are as follows:

1. Make as many as *SMP* copies of the current position of $Cat_k$.
2. For each copy, randomly select as many as *CDC* dimensions to be mutated. Moreover, randomly add or subtract *SRD* values from the current values, which replace the old positions as shown in Equation 1.

$$Xjd_{new} = (1 + rand * SRD) * Xjd_{old} \tag{1}$$

   Where $Xjd_{old}$ is the current position; $Xjd_{new}$ is the next position; *j* denotes the number of a cat and *d* denotes the dimensions; a *rand* is a random number in the interval of [0, 1].

3. Evaluate the fitness value (*FS*) for all the candidate positions.
4. Based on probability select one of the candidate points to be the next position for the cat where candidate points with higher *FS* have more chance to be selected as shown in Equation 2. However, if all fitness values are equal then set all the selecting probability of each candidate point to be 1.

$$Pi = \frac{\left|FS_i - FS_b\right|}{FS_{max} - FS_{min}}, \text{where } 0 < i < j \tag{2}$$

If the objective is minimization then $FS_b = FS_{max}$, otherwise $FS_b = FS_{min}$.

2.3 Tracing Mode: This mode copies the tracing behavior of cats. For the first iteration, random velocity values are given to all dimensions of a cat's position. However, for later steps velocity values need to be updated. Moving cats in this mode are as follows:

1. Update velocities ($V_{k,d}$) for all dimensions according to Equation 3.
2. If a velocity value out-ranged the maximum value, then it is equal to the maximum velocity.

$$V_{k,d} = V_{k,d} + r_1 c_1 (X_{best,d} - X_{k,d}) \tag{3}$$

3. Update position of $Cat_k$ according to Equation 4.

$$X_{k,d} = X_{k,d} + V_{k,d} \tag{4}$$

Refer to (Figure 2) which recaps the whole algorithm in a diagram.

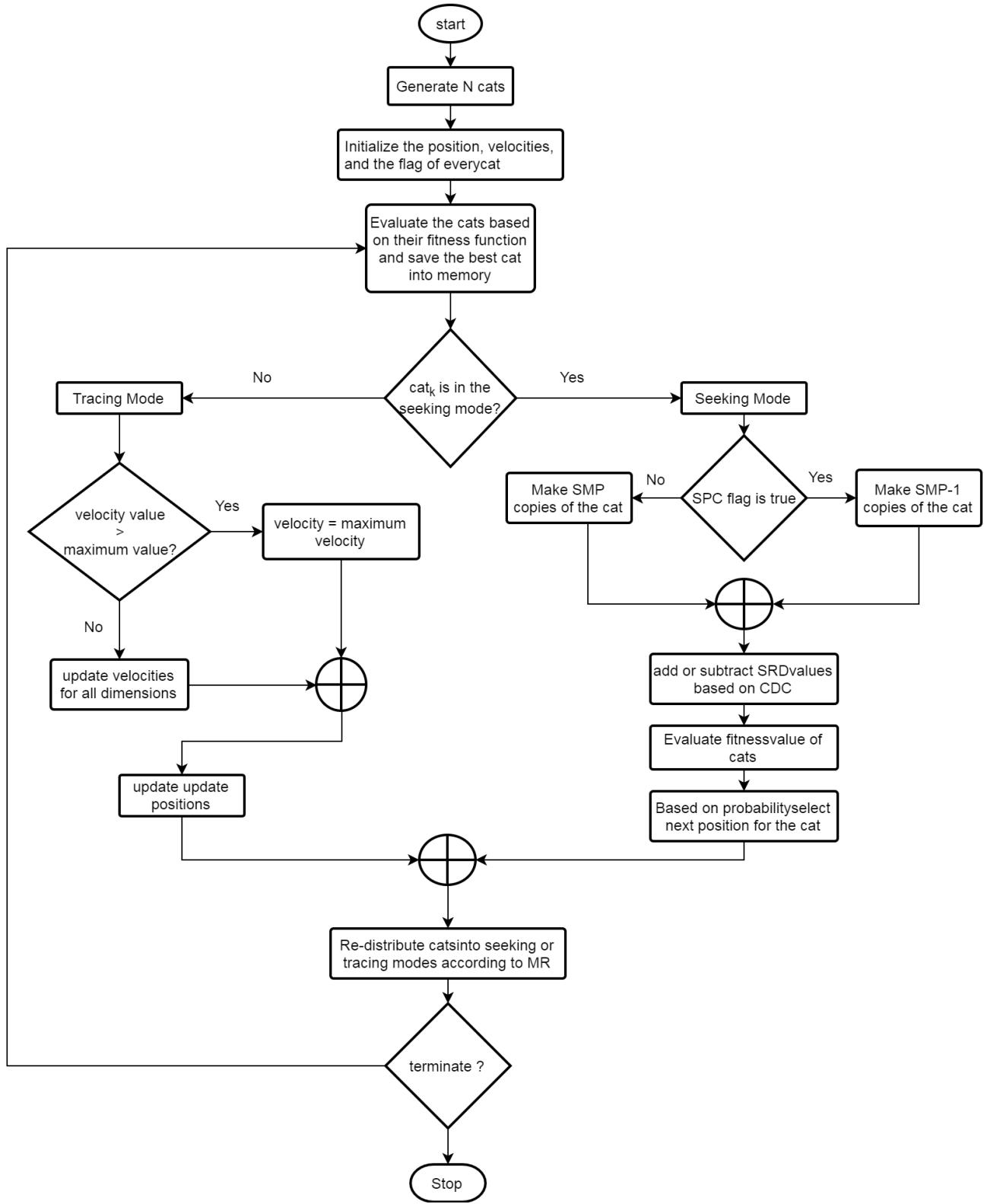

Figure 2: Cat Swarm Optimization algorithm general structure

## **3.** Variants of CSO

In the previous section, the original CSO was covered; this section briefly discusses all other variants of CSO found in the literature. Variants may include the following points: binary or multi-objective versions of the algorithm, changing parameters, altering steps, modifying the structure of the algorithm, or hybridizing it with other algorithms. Refer to (Table 1), which presents a summary of these modifications and their results.

3.1 Discrete Binary Cat Swarm Optimization algorithm (BCSO): Sharafi et al. introduced the BCSO Algorithm, which is the binary version of CSO [9]. In the seeking mode, the SRD parameter has been substituted by another parameter called the probability of mutation operation (PMO). However, the proceeding steps of seeking mode and the other three parameters stay the same. Accordingly, the dimensions are selected using the CDC and then PMO will be applied. In the tracing mode, the calculations of velocity and position equations have also been changed into a new form, in which the new position vector is composed of binary digits taken from either current position vector or global position vector (best position vector). Two velocity vectors are also defined in order to decide which vector (current or global) to choose from.

3.2 Multi-objective Cat Swarm Optimization (MOCSO): Pradhan and Panda proposed multi-objective Cat Swarm Optimization (MOCSO) by extending CSO to deal with multi-objective problems [10]. MOCSO is combined with the concept of the external archive and Pareto dominance in order to handle the non-dominated solutions.

3.3 Parallel Cat Swarm Optimization (PCSO): Tsai and pan introduced Parallel Cat Swarm Optimization (PCSO) [11]. This algorithm improved the CSO algorithm by eliminating the worst solutions. To achieve this, they first distribute the cats into sub-groups i.e. sub-populations. Cats in the seeking mode move as they do in the original algorithm. However, in the tracing mode, for each sub-group, the best cat will be saved into memory and will be considered as the local best. Furthermore, cats move towards the local best rather than the global best. Then, in each group, the cats are sorted according to their fitness function from best to worst. This procedure will continue for a number of iterations, which is specified by a parameter called ECH (a threshold that defines when to exchange the information of groups). For example, if ECH was equal to 20, then once every 20 iterations, the sub-groups exchange information where the worst cats will be replaced by a randomly chosen local best of another group. These modifications lead the algorithm to be computationally faster and show more accuracy when the number of iteration is fewer and the population size is small.

3.4 CSO clustering: Santosa and Ningrum improved the CSO algorithm and applied it for clustering purposes [12]. The main goal was to use CSO to cluster the data and find the best cluster center. The modifications they did were two main points: firstly, removing the mixture ratio (MR) and hence forcing all the cats to go through both seeking and tracing mode. This is aimed at shortening the time required to find the best cluster center; Secondly, always setting the CDC value to be 100%, instead of 80% as in the original CSO, in order to change all dimensions of the candidate cats and increase diversity.

3.5 Enhanced Parallel Cat Swarm Optimization (EPCSO): Tsai et al. further improved the PCSO Algorithm in terms of accuracy and performance by utilizing the orthogonal array of Taguchi method and called it Enhanced Parallel Cat Swarm Optimization (EPCSO) [13]. Taguchi methods are statistical methods, which are invented by Japanese Engineer Genichi Taguchi. The idea is developed based on "ORTHOGONAL ARRAY" experiments, which improves the engineering productivity in the matters of cost, quality, and performance. In their proposed algorithm, the seeking mode of EPCSO is the same as the original CSO. However, the tracing mode has adopted the Taguchi orthogonal array. The aim of this is to improve the computational cost even when the number of agents increases. Therefore, two sets of candidate velocities will be created in the tracing mode. Then, based on the orthogonal array, the experiments will be run and accordingly the position of cats will be updated. [14] Added some partial modifications to EPCSO in order to further improve it and make it fit their application. The modifications were changing the representation of agents from the coordinate to a set; adding a newly defined cluster flag; and designing Custom-Made Fitness Function.

3.6 Average-Inertia Weighted CSO (AICSO): Orouskhani et al. introduced an inertia value to the velocity equation in order to achieve a balance between exploration and exploitation phase. They experimented that (w) value is better to be selected in the range of [0.4, 0.9] where at the beginning of the operation it is set 0.9 and as the iteration number moves forward, (w) value gradually becomes smaller until it reaches 0.4 at the final iteration. Large values of (w) assist global search; whereas small values of (w) assist the local search. In addition to adding inertia value, the position equation was also reformed to a new one, in which averages of current and previous positions, as well as an average of current and previous velocities, were taken in the equation [15].

3.7 Adaptive Dynamic Cat Swarm Optimization (ADCSO): Orouskhani et al. further enhanced the algorithm by introducing three main modifications [16]. Firstly: they introduced an adjustable inertia value to the velocity equation. This value gradually decreases as the dimension numbers increase. Therefore, it has the largest value for dimension one and vice versa. Secondly, they changed the constant (C) to an adjustable value. However, opposite to the inertia weight, it has the smallest value for dimension one and gradually increases until the final dimension where it has the largest value. Finally, they reformed the position equation by taking advantage of other dimensions' information.

3.8 Enhanced Hybrid Cat Swarm Optimization (Enhanced HCSO): Hadi and Sabah proposed a hybrid system and called it: Enhanced HCSO [17,18]. The goal was to decrease the computation cost of the Block matching process in video editing. In their proposal, they utilized a fitness calculation strategy in seeking a mode of the algorithm. The idea was to avoid calculating some areas by deciding whether or not to do the calculation or estimate the next search location to move to. In addition, they also introduced the inertia weight to the tracing mode.

3.9 Improvement Structure of Cat Swarm Optimization (ICSO): Hadi and Sabah proposed combining two concepts together to improve the algorithm and named it ICSO. The first concept is parallel tracing mode and information exchanging, which was taken from PCSO. The second concept is the addition of an inertia weight to the position equation, which was taken from AICSO. They applied their algorithm for Efficient Motion Estimation in block matching. Their goal was to enhance the performance and reduce the number of iterations without the degradation of the image quality [19].

3.10 Opposition-based Learning-Improved CSO (OL-ICSO): Kumar and Sahoo first proposed using Cauchy mutation operator to improve the exploration phase of the CSO algorithm in [20]. Then, they introducing two more modifications to further improved the algorithm and named it: Opposition-based Learning-Improved CSO (OL-ICSO). They improved the population diversity of the algorithm by adopting an opposition-based learning method. Finally, two heuristic mechanisms (for both seeking and tracing mode) were introduced. The goal of introducing these two mechanisms was to improve the diverse nature of the populations and prevent the possibility of falling the algorithm into the local optima when the solution lies near the boundary of the datasets and data vectors cross the boundary constraints frequently [21].

3.11 Chaos Quantum-behaved Cat Swarm Optimization (CQCSO): Nie et al. improved the CSO algorithm in terms of accuracy and avoiding local optima trapping. They first introduced Quantum-behaved Cat Swarm Optimization (QCSO), which combined the CSO algorithm with quantum mechanics. Hence, the accuracy was improved and the algorithm avoided trapping in the local optima. Next, by incorporating a tent map technique, they proposed Chaos Quantum-behaved Cat Swarm Optimization (CQCSO) algorithm. The idea of adding the tent map was to further improve the algorithm and again let the algorithm to jump out of the possible local optima points it might fall into [22].

3.12 Improved Cat Swarm Optimization (ICSO): in the original algorithm, cats are randomly selected to either go into seeking mode or tracing mode using a parameter called MR. However, Kanwar et al. changed the seeking mode by forcing the current best cat in each iteration to move to the seeking mode. Moreover, in their problem domain, the decision variables are firmly integers while solutions in the original cat are continuous. Therefore, from selecting the best cat, two more cats are produced by flooring and ceiling its value. After that, all probable combinations of cats are produced from these two cats [23].

3.13 Improved Cat Swarm Optimization (ICSO): Kumar and Singh made two modifications to the improved CSO algorithm and called it ICSO [24]. They first improved the tracing mode by modifying the velocity and updating position equations. In the velocity equation, a random uniformly distributed vector and two adaptive parameters were added to tune global and local search movements. Secondly, a local search method was combined with the algorithm to prevent local optima problem.

3.14 Hybrid PCSOABC: Tsai et al. proposed a hybrid system by combining PCSO with ABC algorithms and named: Hybrid PCSOABC [25]. The structure simply included running PCSO and ABC consecutively. Since PCSO performs faster with a small population size, the algorithm first, starts with a small population and runs PCSO. After a predefined number of iterations, the population size will be increased and the ABC algorithm starts running. Since the proposed algorithm was simple and did not have any adjustable feedback parameters, it sometimes provided worse solutions than PCSO. Nevertheless, its convergence was faster than PCSO.

3.15 CSO-GA-PSOSVM: Vivek and Reddy proposed a new method by combining CSO with particle swarm intelligence (PSO), Genetic Algorithm (GA), and Support Vector Machine (SVM) and called it CSO-GA-PSOSVM [26]. In their method, they adopted the GA mutation operator into the seeking mode of CSO in order to obtain divergence. In addition, they adopted all GA operators as well as PSO subtraction and addition operators into the tracing mode of CSO in order to obtain convergence. This hybrid meta-heuristic system was then incorporated with the SVM classifier and applied on Facial Emotion Recognition.

3.16 Hybrid CSO Based Algorithm: Skoullis et al. introduced three modifications to the algorithm [27]. Firstly, they combined CSO with a local search refining procedure. Secondly, if the current cat is compared with the global best cat and their fitness value was the same, the global best cat will still be updated by the current cat. The aim of this is to achieve more diversity. Finally, cats are individually selected to go into either seeking mode or tracing mode.

3.17 Hybrid CSO–GA–SA: Sarswat et al. also proposed a hybrid system by combining CSO, GA, and SA and then incorporating it with a modularity based method [28]. They named their algorithm Hybrid CSO-GA-SA. The structure of the system was very simple and straight forward as it was composed of a sequential combination of CSO, GA, and SA. They applied the system to detect overlapping community structures and find near-optimal disjoint communities. Therefore, input datasets were firstly fed into the CSO algorithm for a predefined number of iterations. The resulted cats were then converted into chromosomes and henceforth GA was applied on them. However, GA may fall into local optima and to solve this issue, SA was applied afterward.

3.18 Modified Cat Swarm Optimization (MCSO): Lin et al. combined a mutation operator as a local search procedure with a CSO algorithm to find better solutions in the area of the global best [29]. It is then used to optimize the feature selection and parameters of the support vector machine. Additionally, Mohapatra et al. used the idea of using mutation operation before distributing the cats into seeking or tracing modes [30].

3.19 Normal Mutation Strategy Based Cat Swarm Optimization (NMCSO): Pappula et al. adopted a normal mutation technique to CSO algorithm in order to improve the exploration phase of the algorithm. They used sixteen benchmark functions to evaluate their proposed algorithm against CSO and PSO algorithms [31].

3.20 Improved Cat Swarm Optimization (ICSO): Lin et al. improved the seeking mode of CSO algorithm. Firstly, they used crossover operation to generate candidate positions. Secondly, they changed the value of the new position so that SRD value and current position had no correlations [32]. It is worth mentioning that there are four versions of CSO referenced in [19,23,24,32], all having the same name (ICSO). However, their structures are different.

3.21 Compact Cat Swarm Optimization (CCSO): Zhao M. introduced a compact version of the CSO algorithm. A differential operator was used in the seeking mode of the proposed algorithm to replace the original mutation approach. In addition, a normal probability model was used in order to generate new individuals and denote a population of solutions [33].

3.22 Boolean Binary Cat Swarm Optimization (BBCSO): Siqueira et al. worked on simplifying the binary version of CSO in order to increase its efficiency. They reduced the number of equations, replaced the continues operators with logic gates and finally integrated the roulette wheel approach with the MR parameter [34].

3.23 Hybrid Cat Swarm Optimization - Crow Search Algorithm (CSO-CS): Pratiwi AB. proposed a hybrid system by Combining the CSO algorithm with Crow Search (CS) Algorithm. The algorithm first runs the CSO algorithm followed by the memory update technique of the CS algorithm and then new positions will be generated. She applied her algorithm on Vehicle Routing Problem [35].

Table 1: Summary of the modified versions of the CSO algorithm

| Comparison of | With | Testing Field | Performance | Reference |
|---|---|---|---|---|
| CSO (original) | PSO and weighted-PSO | Six test functions | Better | [4,5] |
| BCSO | GA, BPSO and NBPSO | Four test functions [Sphere, Rastrigin, Ackley, and Rosenbrock] | Better | [9] |
| MOCSO | NSGA-II | Cooperative Spectrum Sensing in Cognitive Radio | Better | [10] |
| PCSO | CSO and weighted-PSO | Three test functions [Rosenbrock, Rastrigin, and Griewank] | Better- when the number of iteration is fewer and the population size is small | [11] |
| CSO clustering | K-means and PSO clustering | Four different clustering datasets [Iris, Soybean, Glass and Balance Scale] | More accurate but slower. | [12] |
| EPCSO | PCSO, PSO-LDIW, PSO-CREV, GCPSO, MPSO-TVAC, CPSO-H6, PSO-DVM | Five test functions and aircraft schedule recovery problem | Better | [13] |
| AICSO | CSO | Three test function [Rastrigin, Griewank, and Ackley] | Better | [15] |
| ADCSO | CSO | Six test functions [Rastrigin, Griewank, Ackley, Axis parallel, Trid10, and Zakharov] | better - except for Griewank test function. | [16] |
| Enhanced HCSO | PSO | Motion estimation block-matching | Better | [17,18] |
| ICSO | PSO | Motion estimation block-matching | Better | [19] |
| OL-ICSO | K-Median, PSO, CSO, and ICSO | ART1, ART2, iris, CMC, cancer, and wine datasets | Better | [21] |
| CQCSO | QCSO, CSO, PSO, and CPSO | Five test functions [Schaffer, Shubert, Griewank, Rastrigin, and Rosenbrock] and multipeak maximum power point tracking for a photovoltaic array under complex conditions | Better | [22] |
| ICSO | CSO and PSO | The 69-bus test distribution system | Better | [23] |
| ICSO | CSO, BCSO, AICSO, and EPCSO | Twelve test functions [Sphere, Rosenbrock, Rastrigin, Griewank, Ackley, Step, Powell, Schwefel, Schaffer, Zakharov's, Michalewicz, Quartic] and five real-life clustering problems [iris, cancer, CMC, wine and glass] | Better | [24] |
| Hybrid PCSOABC | PCSO and ABC | Five test functions | Better | [25] |
| CSO-GA-PSO$_{SVM}$ | CSO+SVM (CSO$_{SVM}$) | 66 feature points from each face of CK+ (Cohn Kanade) dataset | better | [26] |
| Hybrid CSO Based Algorithm | GA, EA, SA, PSO, and AFS | school timetabling test instances | better | [27] |

| Hybrid CSO-GA-SA | SLPA and CFinder | seven datasets [Karate, Dolphin, Polbooks, Football, Net-Science, Power, Indian Railway] | better | [28] |
|---|---|---|---|---|
| MCSO | CSO | Nine datasets from UCI | better | [29] |
| MCSO | CSO | Eight dataset | better | [30] |
| NMCSO | CSO, PSO | Sixteen benchmark function | better | [31] |
| ICSO | CSO | Ten datasets from UCI | better | [32] |
| cCSO | DE, PSO, CSO | 47 benchmark functions | better | [33] |
| BBCSO | Binary Particle Swarm Optimization (BPSO), Binary Genetic Algorithm (BGA), Binary CSO | 0/1 Knapsack Optimization problem | better | [34] |
| CSO-CS | N/A | VRP instances from http://neo.lcc.uma.es/vrp/ | N/A | [35] |

## 4. CSO and its variants with Artificial Neural Networks

Artificial Neural Networks are computing systems, which have countless numbers of applications in various fields. Earlier Neural Networks used to be trained by conventional methods, such as the Back Propagation algorithm. However, current Neural Networks are trained by Nature-inspired optimization algorithms. The training could be optimizing the node weights or even the network architectures [36]. CSO has also been extensively combined with Neural Networks in order to be applied in different application areas. This section briefly goes over those works, in which CSO is hybridized with ANN and similar methods.

4.1 CSO + ANN + OBD: Yusiong proposes combining ANN with CSO algorithm and Optimal Brain Damage (OBD) approach. Firstly, the CSO algorithm is used as an optimization technique to train the ANN algorithm. Secondly, OBD is used as a pruning algorithm to decrease the complexity of ANN structure where less number of connections has been used. As a result, an Artificial Neural Network was obtained that had less training errors and high classification accuracy [37].

4.2 ADCSO+GD+ANFIS: Orouskhani et al. combined the ADCSO algorithm with the Gradient Descent Algorithm (GD) in order to tweak parameters of the Adaptive Network-Based Fuzzy Inference System (ANFIS). In their method, the antecedent and consequent parameters of ANFIS were trained by the CSO algorithm and the GD algorithm consecutively [38].

4.3 CSO+SVM: Abed and Alasadi proposed a hybrid system based on SVM and CSO. The system was applied to Electrocardiograms Signals classification. They used CSO for the purpose of feature selection optimization and enhancing SVM parameters[39]. In addition, [40,41] also combined CSO with SVM and applied it to a Classroom Response System.

4.4 CSO+WNN: Nanda proposed a hybrid system by combining Wavelet Neural Network (WNN) and CSO algorithm. In their proposal, the CSO algorithm was used to train the weights of WNN in order to obtain the near-optimal weights [42].

4.5 BCSO+SVM: Mohamadeen et al. built a classification model based on BCSO and SVM and then applied it in a power system. The use of BCSO was to optimize SVM parameters [43].

4.6 CCSO+ANN: Wang et al. proposed designing an ANN that can handle randomness, fuzziness, and accumulative time effect in time series concurrently. In their work, the CSO algorithm was used to optimize the network structure and learning parameters at the same time [44].

4.7 CSO/PSO+ANN: Chittineni et al. used CSO and PSO algorithms to train ANN and then applied their method on stock market prediction. Their comparison results showed that the CSO algorithm performed better than the PSO algorithm. [45]

4.8 CS-FLANN: Kumar et al. combined the CSO algorithm with Functional Link Artificial Neural Network (FLANN) to develop an evolutionary filter to remove Gaussian noise [46].

## 5. Applications of CSO

This section presents the applications of the CSO algorithm, which are categorized into six groups namely, Electrical Engineering, Computer Vision, Signal Processing, System Management, and Combinatorial Optimization, Wireless and WSN, Petroleum Engineering and Civil Engineering. A summary of the purposes and results of these applications is provided in (Table 2).

5.1 Electrical Engineering: CSO algorithm has been extensively applied in the electrical engineering field. Hwang et al. applied both CSO and PSO algorithms on an electrical payment system in order to minimize electricity costs for customers. Results indicated that CSO is more efficient and faster than PSO in finding the global best solution [47]. Economic Load Dispatch (ELD) and Unit Commitment (UC) are significant applications, in which the goal is to reduce the total cost of fuel is a power system. Chen et al. applied the CSO algorithm on the Economic Load Dispatch (ELD) of wind and thermal generators [48]. Faraji et al. also proposed applying the Binary Cat swarm Optimization (BCSO) algorithm on UC and obtained better results compared to the previous approaches [49]. UPFC stands for the Unified Power Flow Controller, which is an electrical device used in transmission systems to control both active and reactive power flows. Kumar, G.N. and M.S. Kalavathi used the CSO algorithm to optimize UPFC in order to improve the stability of the system [50]. Lenin, K. and B.R. Reddy also applied ADCSO on reactive power dispatch problem in the aim to minimize active power loss [51]. Improving Available Transfer Capability (ATC) is very significant in electrical engineering. Nireekshana, T., G.K. Rao, and S.S. Raju used the CSO algorithm to regulate the position and control parameters of SVC and TCSC in the aim of Maximizing power transfer transactions during normal and contingency cases [52]. The function of the transformers is to deliver electricity to consumers. Determining how reliable these transformers are in a power system is essential. Mohamadeen, K., R.M. Sharkawy, and M. Salama proposed a classification model to classify the transformers according to their reliability status [43]. The model was built based on BCSO incorporation with SVM. The results are then compared with a similar model based on BPSO. It is shown that BCSO is more efficient in optimizing the SVM parameters. Wang et al. proposed designing an ANN that can handle randomness, fuzziness, and accumulative time effect in time series concurrently [44]. In their work, the CSO algorithm has been used to optimize the network structure and learning parameters at the same time. Then, the model was applied to two applications, which were individual household electric power consumption forecasting and Alkaline-suthe rfactant-polymer(ASP) flooding oil recovery index forecasting in oilfield development. The Current Source Inverter (CSI) is a conventional kind of power inverter topologies. Hosseinnia and Farsadi combined Selective Harmonic Elimination (SHE) in corporation with CSO algorithm and then applied it on Current Source Inverter (CSI) [53]. The role of the CSO algorithm was to optimize and tune the switching parameters and minimize total harmonic distortion. [54] used CSO and PCSO to find the optimal place and size of distributed generation units on distribution networks. [55] used MCSO algorithm to propose a novel maximum power point tracking (MPPT) approach to obtain global maximum power point (GMPP) tracking. Srivastava et al. used BCSO algorithm to optimize the location of phasor measurement units and reduce the required number of PMUs [56]. Guo L et al. used CSO algorithm to identify the parameters of single and double diode models in solar cell models [57].

5.2 Computer vision: Facial Emotion Recognition is a biometric approach to identify human emotion and classify them accordingly. References [40,41] proposed a classroom response system by combining the CSO algorithm with a support vector machine to classify student's facial expressions. Vivek, T. and G.R.M. Reddy also used the CSO-GA-PSOSVM algorithm for the same purpose [26]. Block matching in video processing is computationally expensive and time-consuming. Hadi, I. and M. Sabah used the CSO algorithm in block matching for efficient motion estimation [58]. The aim was to decrease the number of positions that needs to be calculated within the search window during the block matching process i.e. to enhance the performance and reduce the number of iterations without the degradation of the image quality. The authors further improved their work and achieved better results by replacing the CSO algorithm with HCSO and ICSO in [17,18] respectively. References [59,60] used CSO Algorithm to retrieve watermarks similar to the original copy. In video processing, object tracking is the process of determining the position of a moving object over time using a camera. Hadi, I. and M. Sabah used EHCSO in an object-tracking system for further enhancement in terms of efficiency and accuracy [61]. Yan, L., X. Yan-Qiu, and W. Li-Hai used BCSO as a band selection method for hyperspectral images [62]. In computer vision, image segmentation refers to the process of dividing an image into multiple parts. Reference [63,64] proposed using CSO algorithm incorporation with the concept of multilevel thresholding for image segmentation purposes. Zhang et al. combined wavelet entropy, ANN, and CSO algorithm to develop an Alcohol Use Disorder (AUD) identification system [65]. Kumar et al. combined the CSO algorithm with Functional Link Artificial Neural Network (FLANN) to remove the unwanted Gaussian Noise from CT images [46]. Yang et al. combined CSO with L-BFGS-B technique to register non-rigid multi-modal images [66]. Ilhan and Aydin employed the CSO algorithm to tune the parameters in the histogram stretching technique for the purpose of image enhancement [67].

5.3 Signal processing: IIR filter stands for Infinite impulse response. It is a discrete-time filter, which has applications in signal processing and communication. Panda, G., P.M. Pradhan, and B. Majhi used the CSO algorithm for IIR system identification [68]. The authors also applied the CSO algorithm as an optimization mechanism to do direct and inverse modeling of linear and nonlinear plants [69]. Abed, M.A., and H.A.A. Alasadi combined CSO Algorithm with SVM for Electrocardiograms Signal Classification [39]

5.4 System management and combinatorial optimization: In parallel computing, optimal task allocation is a key challenge. [70,71] proposed using the CSO algorithm to maximize system reliability. There are three basic scheduling problems, namely open shop, job shop, and flow shop. These problems are classified as NP-hard and have many real-world applications. They coordinate assigning jobs to resources at particular times, where the objective is to minimize time consumption. However, their difference is mainly in having ordering constraints on operations. Bouzidi et al. applied the BCSO algorithm on the job scheduling problem (JSSP) in [72]. They also made a comparative study between CSO and two other meta-heuristic algorithms namely: Cuckoo search algorithm (CS), and the Ant Colony Optimization (ACO) for JSSP in [73]. Then, they used the CSO algorithm to solve Flow shop scheduling (FSSP) [74] and open shop scheduling problems (OSSP) as well [75]. Moreover, Dani et al. also applied the CSO algorithm on JSSP in which they used a non-conventional approach to represent cat positions [76]. Maurya and Tripathi also applied the CSO algorithm on Bag-of-tasks and workflow scheduling problems in cloud systems [77]. Bouzidi, A. and M.E. Riffi applied CSO Algorithm on the Traveling Salesman Problem (TSP) and the Quadratic Assignment Problem (QAP), which are two combinatorial optimization problems [78]. Bouzidi et al. also made a comparative study between CSO algorithm, cuckoo search algorithm, and bat-inspired algorithm for addressing TSP [79]. In cloud computing, minimizing the total execution cost while allocating tasks to processing resources is a key problem. Bilgaiyan, S., S. Sagnika, and M. Das applied CSO and MCSO algorithms on workflow scheduling in cloud systems [80]. In addition, Kumar et al also applied BCSO on workflow scheduling in Cloud systems [81]. Set Cover Problem (SCP) is considered as an NP-complete problem. Crawford et al. successfully applied the BCSO Algorithm to this problem [82]. They further improved this work by using Binarization techniques and selecting different parameters for each test example sets [83,84]. Web Services provide a standardized communication between applications over the web which have many important applications. However, discovering appropriate web services for a given task is challenging. Kotekar, S. and S.S. Kamath used a CSO based approach as a clustering algorithm to group service documents according to their functionality similarities [85]. Sarswat, A., V. Jami, and R.M.R. Guddeti applied Hybrid CSO–GA–SA to detect the overlapping community structures and find the near-optimal disjoint communities [28]. Optimizing the problem of controlling complex network systems is critical in many areas of science and engineering. Orouskhani, Y., M. Jalili, and X. Yu apply CSO algorithm to address a number of problems in optimal pinning controllability and thus optimize the network structure [86]. Skoullis et al. combined the CSO algorithm with a local search refining procedure and applied it on high school timetabling problem [27]. Soto et al. combined BCSO with Dynamic mixture ratios to organize the cells in Manufacturing cell Design Problem [87]. Bahrami et al. applied a CSO algorithm on water resource management where the algorithm was used to find the optimal Reservoir Operation [88]. Kencana et al. used CSO algorithm to classify the feasibility of small loans in banking systems [89]. Majumder et al. combined the CSO algorithm with the analytic element method (AEM) and reverse particle tracking (RPT) to model novel Groundwater Management systems [90]. Rautray et al. used CSO algorithm to solve the multi-document summarization problem [91]. Thomas et al. combined radial point collocation meshfree (RPCM) approach with CSO algorithm to be used in the groundwater resource management [92]. Pratiwi created a hybrid system by combining the CSO algorithm and Crow Search (CS) Algorithm and then used it to address the Vehicle Routing Problem with time windows (VRPTW) [93]. Naem et al. proposed a modularity based system by combining the CSO algorithm with K-median clustering technique to detect overlapping community in social networks [94].

5.5 Wireless and WSN: The ever-growing wireless devices push researchers to use electromagnetic spectrum bands more wisely. Cognitive Radio (CR) is an effective dynamic spectrum allocation in which spectrums are dynamically assigned based on a specific time or location. Pradhan, P.M. and G. Panda in [95,96] combined MOCSO with fitness sharing and fuzzy mechanism and applied it on CR design. they also conducted a comparative analysis and proposed a generalized method to design a CR engine based on six evolutionary algorithms [97]. Wireless Sensor Network (WSN) refers to a group of nodes (wireless sensors) that form a network to monitor physical or environmental conditions. The gathered data need to be forwarded among the nodes and each node requires having a routing path. Kong et al. proposed applying Enhanced Parallel Cat Swarm Optimization (EPCSO) algorithm in this area as a routing algorithm [14]. Another concern in the context of WSN is minimizing the total power consumption while satisfying the performance criterions. So, Tsiflikiotis, A. and S.K. Goudos addressed this problem which is known as optimal power allocation problem, and for that three meta-heuristic algorithms were presented and compared [98]. Moreover, Pushpalatha, A. and G. Kousalya applied CSO in WSN for optimizing cluster head selection which helps in energy saving and available bandwidth [99]. Alam et al. also applied the CSO algorithm in a clustering-based method to handle Channel Allocation (CA) issues between secondary users with respect to practical constraints in the Smart Grid environment [100,101,102] used the CSO algorithm to find the optimal location of sink nodes in W SN. Ram et al. applied CSO algorithm to minimize the sidelobe level of antenna arrays and enhance the Directivity [103]. Ram et al. used CSO to optimize controlling parameters of linear antenna arrays and produce optimal designs [104]. Pappula et al. also used Cauchy mutated CSO to make linear aperiodic arrays, where the goal was to reduce sidelobe level and control the null positions [105].

5.6 Petroleum Engineering: CSO algorithm has also been applied in the petroleum engineering field. For example, it was used as a good placement optimization approach by Chen et al. in [106,107]. Furthermore, Wang et al. used the CSO algorithm as an ASP flooding oil recovery index forecasting approach [44].

5.7 Civil Engineering: Ghadim et al. used the CSO algorithm to create an identification model that detects early cracks in building structures [108].

Table 2: the purposes and results of using CSO algorithm in various applications

| Purpose | Results | Ref. |
|---|---|---|
| CSO applied on electrical payment system in order to minimize electricity cost for customers | CSO outperformed PSO | [47] |
| CSO applied on Economic Load Dispatch (ELD) of wind and thermal generator | CSO outperformed PSO | [48] |
| BCSO applied on Unit Commitment (UC) | CSO outperformed LR, ICGA, BF, MILP, ICA, and SFLA | [49] |
| Applied CSO algorithm on UPFC to increase the stability of the system | IEEE 6 bus and 14 bus networks were used in the simulation experiments  and desirable results were achieved | [50] |
| Applied ADCSO on reactive power dispatch problem to minimize active power loss | IEEE 57-bus system was used in the simulation experiments, in which ADCSO outperformed 16 other optimization algorithms | [51] |
| Applied CSO algorithm to regulate the position and control parameters of SVC and TCSC to improve Available Transfer Capability (ATC) | IEEE 14-bus and IEEE 24-bus systems were used in the simulation experiments, in which the system provided better results after adopting CSO | [52] |
| Building a classification model based on BCSO and SVM to classify the transformers according to their reliability status. | The model performed better compared to a similar model, which was based on BPSO and VSM | [43] |
| Applied CSO to optimize the network structure and learning parameters of an ANN model named (CPNN-CSO), which is used to predict household electric power consumption | CPNN-CSO outperformed ANFIS and similar methods with no CSO such as PNN and CPNN | [44] |
| Applied CSO and Selective Harmonic Elimination (SHE) algorithm on Current Source Inverter (CSI) | CSO successfully optimized the switching parameters of CSI and hence minimized the total harmonic distortion | [53] |
| Applied both CSO, PCSO, PSO–CFA, and ACO–ABC on distributed generation units on distribution networks | IEEE 33-bus and IEEE 69-bus distribution systems were used in the simulation experiments and CSO outperformed the other algorithms | [54] |
| Applied MCSO on MPPT to achieve global maximum power point (GMPP) tracking | MCSO outperformed PSO, MPSO, DE, GA and HC algorithms | [55] |
| Applied BCSO to optimize the location of phasor measurement units and reduce the required number of PMUs | IEEE 14-bus and IEEE 30-bus test system was used in the simulation. BCSO outperformed BPSO, Generalized Integer Linear Programming, and Effective Data Structure Based Algorithm | [56] |
| used CSO algorithm to identify the parameters of single and double diode models in the solar cell system | CSO outperformed PSO, GA, SA, PS, Newton, HS, GGHS, IGHS, ABSO, DE, and LMSA | [57] |
| Applied CSO and SVM to classify students' facial expression | The results show 100% classification accuracy for the selected 9 face expressions | [40] |
| Applied CSO and SVM to classify students' facial expression | The system achieved satisfactory results | [41] |
| Applied CSO-GA-PSOSVM to classify students' facial expression | The system achieved 99% classification accuracy | [26] |
| Applied CSO, HCSO, and ICSO in block matching for efficient motion estimation | The system reduced computational complexity and provided faster convergence | [58,17,18] |
| Used CSO Algorithm to retrieve watermarks similar to the original copy | CSO outperformed PSO and PSO time-varying inertia weight factor algorithms | [59,60] |
| Sabah used EHCSO in an object-tracking system to obtain further efficiency and accuracy | The system yielded desirable results in terms of efficiency and accuracy | [61] |
| used BCSO as a band selection method for hyperspectral images | BCSO outperformed PSO | [62] |
| Used CSO and multilevel thresholding for image segmentation | CSO outperformed PSO | [63] |
| Used CSO and multilevel thresholding for image segmentation | PSO outperformed CSO | [64] |
| Used CSO, ANN and wavelet entropy to build an AUD identification system. | CSO outperformed GA, IGA, PSO, and CSPSO | [65] |
| Used CSO and FLANN to remove the unwanted Gaussian Noises from CT images | The proposed system outperformed Mean Filter and Adaptive Wiener Filter. | [46] |
| Used CSO with L-BFGS-B technique to register non-rigid multi-modal images | The system yielded satisfactory results | [66] |
| Used CSO in image enhancement to optimize parameters of the histogram stretching technique | PSO outperformed CSO | [67] |
| Used CSO algorithm for IIR system identification | CSO outperformed GA and PSO | [68] |
| Applied CSO to do direct and inverse modeling of linear and nonlinear plants | CSO outperformed GA and PSO | [69] |
| Used CSO and SVM for Electrocardiograms Signal Classification | Optimizing SVM parameters using CSO improved the system in terms of accuracy | [39] |
| Applied CSO to increase reliability in a task allocation system | CSO outperformed GA and PSO | [70,71] |
| Applied CSO on JSSP | The benchmark instances were taken from OR-Library. CSO yielded desirable results compared to the best recorded results in the dataset reference. | [72] |
| Applied BCSO on JSSP | ACO outperformed CSO and Cuckoo Search algorithms | [73] |
| Applied CSO on FSSP | Carlier, Heller, and Reeves benchmark instances were used, CSO can solve problems of up to 50 jobs accurately | [74] |
| Applied CSO on OSSP | CSO performs better than six Metaheuristic algorithms in the literature. | [75] |
| Applied CSO on JSSP | CSO performs better than some conventional algorithms in terms of accuracy and speed. | [76] |
| Applied CSO on Bag-of-tasks and workflow scheduling problems in cloud systems | CSO performs better than PSO and two other heuristic algorithms | [77] |
| applied CSO on TSP and QAP | The benchmark instances were taken from TSPLIB and QAPLIB. The results show that CSO outperformed the best results recorded in those dataset references. | [78] |
| Comparison between CSO, chukoo search and bat-inspired algorithm to solve TSP problem | The benchmark instances are taken from STPLIB. The results show that CSO falls behind the other algorithms | [79] |
| applied CSO and MCSO on workflow scheduling in cloud systems | CSO performs better than PSO | [80] |
| Applied BCSO on workflow scheduling in Cloud systems | BCSO performs better than PSO and BPSO | [81] |
| Applied BCSO on SCP | BCSO performs better than ABC | [82] |
| Applied BCSO on SCP | BCSO performs better than Binary Teaching-Learning-Based Optimization (BTLBO) | [83,84] |
| Used a CSO as a clustering mechanism in web services. | CSO performs better than K-means | [85] |
| Applied Hybrid CSO–GA–SA to find overlapping community structures. | Very good results were achieved. Silhouette coefficient was used to verify these results in which was between 0.7-0.9 | [28] |
| Used CSO to optimize the network structures for pinning control | CSO outperformed a number of heuristic methods | [86] |
| Applied CSO with local search refining procedure to address high school timetabling problem | CSO outperformed the Genetic Algorithm (GA), Evolutionary Algorithm (EA), Simulated Annealing (SA), Particle Swarm Optimization (PSO) and Artificial Fish Swarm (AFS). | [27] |
| BCSO with Dynamic mixture ratios to address the Manufacturing Cell Design Problem | BCSO can effectively tackle the MCDP problem regardless of the scale of the problem | [87] |
| used CSO to find the optimal Reservoir Operation in water resource management | CSO outperformed GA | [88] |
| Applied CSO to classify the the feasibility of small loans in banking systems | CSO resulted in 76% of accuracy in comparison to 64% resulted from OLR procedure. | [89] |
| Used CSO, AEM, and RPT to build a Groundwater Management systems | CSO outperformed a number of metaheuristic algorithms in addressing groundwater management problem | [90] |
| Applied CSO to solve the multi-document summarization problem | CSO outperformed Harmonic Search (HS) and PSO | [91] |
| Used CSO and (RPCM) to address groundwater resource management | CSO outperformed a similar model based on PSO | [92] |
| applied CSO-CS to solve VRPTW | CSO-CS successfully solves the VRPTW problem. The results show that the algorithm convergences faster by increasing the population and decreasing the $cdc$ parameter. | [93] |
| Applied CSO and K-median to detect overlapping community in social networks | CSO and K-median provides better modularity than similar models based on PSO and BAT algorithm | [94] |
| Applied MOCSO, fitness sharing and fuzzy mechanism on CR design | MOCSO outperformed MOPSO, NSGA-II, and MOBFO | [95,96] |
| Applied CSO and five other metaheuristic algorithms to design a CR engine | CSO outperformed the GA, PSO, DE, BFO and ABC algorithms | [97] |
| Applied EPCSO on WSN to be used as a routing algorithm | EPCSO outperformed AODV, a ladder diffusion using ACO and a ladder diffusion using CSO. | [14] |
| Applied CSO on WSN in order to solve the optimal power allocation problem | PSO is marginally better for small networks. However, CSO outperformed PSO and Chukoo search algorithm | [98] |
| Applied CSO on WSN to optimize cluster head selection | The proposed system outperformed the existing systems by 75%. | [99] |
| Applied CSO on CR based Smart Grid communication network to optimize channel allocation | The proposed system obtains desirable results for both fairness-based and priority-based cases | [100] |
| Applied CSO in WSN to detect the optimal location of sink nodes | CSO outperformed PSO in reducing total power consumption. | [101,102] |
| Applied CSO on Time Modulated Concentric Circular Antenna Array to to minimize the sidelobe level of antenna arrays and enhance the Directivity | CSO outperformed RGA, PSO and DE algorithms | [103] |
| Applied CSO to optimize the radiation pattern controlling parameters for linear antenna arrays. | CSO successfully tunes the parameters and provides optimal designs of linear antenna arrays. | [104] |
| Applied Cauchy mutated CSO to make linear aperiodic arrays, where the goal was to reduce sidelobe level and control the null positions | The proposed system outperformed both CSO and PSO | [105] |
| Applied CSO and Analytical formula-based objective function to optimize well placements | CSO outperformed DE algorithm | [106] |
| Applied CSO to optimize well placements considering oilfield constraints during development. | CSO outperformed GA and DE algorithms | [107] |
| CSO applied to optimize the network structure and learning parameters of an ANN model, which is used to predict an ASP flooding oil recovery index | The system successfully forecast the ASP flooding oil recovery index | [43] |
| Applied CSO to build an identification model to detect early cracks in beam type structures | CSO yields a desirable accuracy in detecting early cracks | [108] |

## 6. Performance Evaluation:

Many variants and applications of the CSO algorithm were discussed in the above sections. However, benchmarking these versions and conducting a comparative analysis between them was not feasible in this work. This is because: firstly, their source codes were not available. Secondly, different test functions or datasets have been used during their experiments. In addition, since the emerging CSO algorithm, many novel and powerful meta-heuristic algorithms have been introduced. However, the literature lacks a comparative study between the CSO algorithm and these new algorithms. Therefore, we conducted an experiment, in which the original CSO algorithm was compared against three new and robust algorithms, which were Dragonfly Algorithm (DA) [6], Butterfly Optimization Algorithm (BOA) [7] and Fitness Dependent Optimizer (FDO) [8]. For this, 23 traditional and 10 modern benchmark functions were used. (See Figure 3), which illustrates the general framework for conducting the performance evaluation process. It is worth mentioning that for four test functions, BOA returned imaginary numbers and we set "N/A" for them.

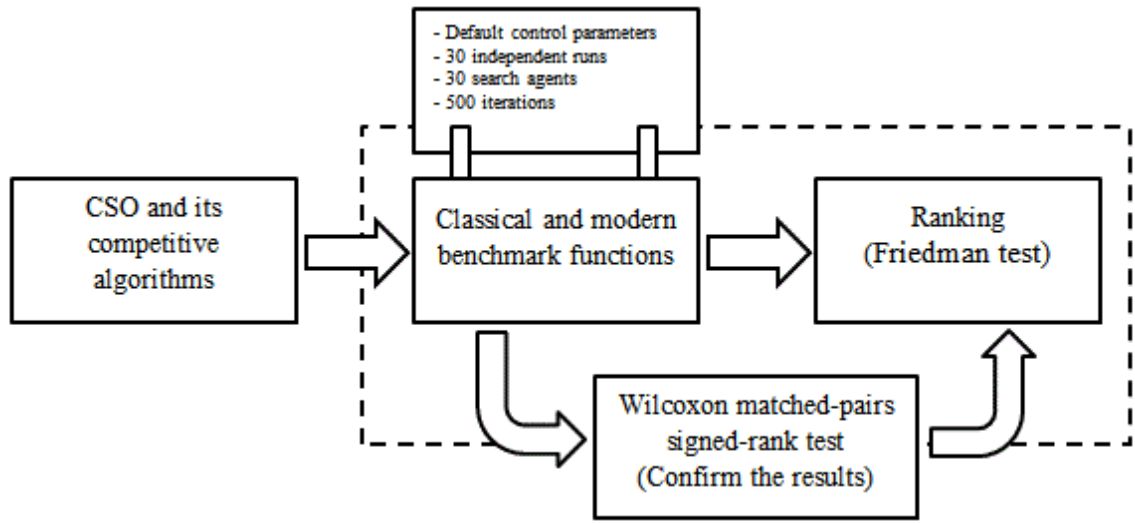

Figure 3: General framework of the performance evaluation process

6.1 Traditional benchmark functions: This group includes the unimodal and multimodal test functions. Unimodal test functions contain one single optimum while, multimodal test functions contain multiple local optima and usually a single global optimum. F1 to F7 are unimodal test functions (Table (3)), which are employed to experiment with the global search capability of the algorithms. Furthermore, F8 to F23 are multimodal test functions, which are employed to experiment with the local search capability of the algorithms. Refer to ref [109] for the detailed description of unimodal and multimodal functions.

6.2 Modern benchmark functions (CEC 2019): These set of benchmark functions, also called composite benchmark functions, are complex and difficult to solve. The CEC01 to CEC10 functions as shown in Table (3) are of these types, which are shifted, rotated, expanded, and combined versions of traditional benchmark functions. Refer to ref [110] for the detailed description of modern benchmark functions.

The comparison results for CSO and other algorithms are given in Table (3) in the form of mean and standard deviations. For each test function, the algorithms are executed 30 independent runs. For each run, 30 search agents were searching over the course of 500 iterations. Parameter settings are set as defaults for all algorithms and nothing was changed.

Table 3: Comparison results of CSO algorithm with modern Meta-heuristic Algorithms

| Functions | CSO | | DA | | BOA | | FDO | | $f_{min}$ |
|---|---|---|---|---|---|---|---|---|---|
| | AV | STD | AV | STD | AV | STD | AV | STD | |
| F1 | 3.50E-14 | 6.34E-14 | 15.24805 | 23.78914 | 1.01E-11 | 1.66E-12 | 2.13E-23 | 1.06E-22 | 0 |
| F2 | 2.68E-08 | 2.61E-08 | 1.458012 | 0.869819 | 4.65E-09 | 4.63E-10 | 0.047175 | 0.188922 | 0 |
| F3 | 7.17E-09 | 1.16E-08 | 136.259 | 151.9406 | 1.08E-11 | 1.71E-12 | 2.39E-06 | 1.28E-05 | 0 |
| F4 | 0.010352 | 0.007956 | 3.262584 | 2.112636 | 5.25E-09 | 5.53E-10 | 4.93E-08 | 9.09E-08 | 0 |
| F5 | 8.587858 | 0.598892 | 374.9048 | 691.5889 | 8.935518 | 0.02146 | 21.58376 | 39.66721 | 0 |
| F6 | 1.151759 | 0.431511 | 12.07847 | 17.97414 | 1.04685 | 0.346543 | 7.15E-22 | 2.80E-21 | 0 |
| F7 | 0.026026 | 0.015039 | 0.035679 | 0.023538 | 0.001513 | 0.00056 | 0.612389 | 0.299315 | 0 |
| F8 | -2855.11 | 359.1697 | -2814.14 | 432.944 | NA | NA | -10502.1 | 15188.77 | -418.9829 x 5 |
| F9 | 24.01772 | 6.480946 | 26.53478 | 11.20011 | 28.6796 | 20.17813 | 7.940883 | 4.110302 | 0 |
| F10 | 3.754226 | 1.680534 | 2.827344 | 1.042434 | 3.00E-09 | 1.16E-09 | 7.76E-15 | 2.46E-15 | 0 |
| F11 | 0.355631 | 0.19145 | 0.680359 | 0.353454 | 1.35E-13 | 6.27E-14 | 0.175694 | 0.148586 | 0 |
| F12 | 1.900773 | 1.379549 | 2.083215 | 1.436402 | 0.130733 | 0.084891 | 7.737715 | 4.714534 | 0 |
| F13 | 1.160662 | 0.53832 | 1.072302 | 1.327413 | 0.451355 | 0.138253 | 4.724571 | 6.448214 | 0 |
| F14 | 0.998004 | 3.39E-07 | 1.064272 | 0.252193 | 1.52699 | 0.841504 | 2.448453 | 1.766953 | 1 |
| F15 | 0.001079 | 0.00117 | 0.005567 | 0.012211 | 0.000427 | 9.87E-05 | 0.001492 | 0.003609 | 0.00030 |
| F16 | -1.03162 | 1.53E-05 | -1.03163 | 4.76E-07 | NA | NA | -1.00442 | 0.149011 | -1.0316 |
| F17 | 0.304253 | 1.81E-06 | 0.304251 | 0 | 0.310807 | 0.004984 | 0.397887 | 5.17E-15 | 0.398 |
| F18 | 3.003667 | 0.004338 | 3.000003 | 1.22E-05 | 3.126995 | 0.211554 | 3 | 2.37E-07 | 3 |
| F19 | -3.8625 | 0.00063 | -3.86262 | 0.00037 | NA | NA | -3.86015 | 0.003777 | -3.86 |
| F20 | -3.30564 | 0.045254 | -3.25226 | 0.069341 | NA | NA | -3.06154 | 0.380813 | -3.32 |
| F21 | -9.88163 | 0.90859 | -7.28362 | 2.790655 | -4.44409 | 0.383552 | -4.19074 | 2.664305 | -10.1532 |
| F22 | -10.2995 | 0.094999 | -8.37454 | 2.726577 | -4.1496 | 0.715469 | -4.89633 | 3.085016 | -10.4028 |
| F23 | -10.0356 | 1.375583 | -6.40669 | 2.892797 | -4.12367 | 0.859409 | -4.03276 | 2.517357 | -10.5363 |
| CEC01 | 1.58E+09 | 1.71E+09 | 3.8E+10 | 4.03E+10 | 58930.69 | 11445.72 | 4585.278 | 20707.63 | 1 |
| CEC02 | 19.70367 | 0.580672 | 83.73248 | 100.1326 | 18.91597 | 0.291311 | 4 | 3.28E-09 | 1 |
| CEC03 | 13.70241 | 25.3E-06 | 13.70263 | 0.000673 | 13.70321 | 0.000617 | 13.7024 | 1.68E-11 | 1 |
| CEC04 | 179.1984 | 55.37322 | 371.2471 | 420.2062 | 20941.5 | 7707.688 | 33.08378 | 16.81143 | 1 |
| CEC05 | 2.671378 | 0.171923 | 2.571134 | 0.304055 | 6.176949 | 0.708134 | 2.13924 | 0.087218 | 1 |
| CEC06 | 11.21251 | 0.708359 | 10.34469 | 1.335367 | 11.83069 | 0.771166 | 12.13326 | 0.610499 | 1 |
| CEC07 | 365.2358 | 164.997 | 534.3862 | 240.0417 | 1043.895 | 215.3575 | 120.4858 | 13.82608 | 1 |
| CEC08 | 5.499615 | 0.484645 | 5.86374 | 0.51577 | 6.337199 | 0.359203 | 6.102152 | 0.769938 | 1 |
| CEC09 | 6.325862 | 1.295848 | 8.501541 | 16.90603 | 2270.616 | 811.4442 | 2 | 2.00E-10 | 1 |
| CEC10 | 21.36829 | 0.06897 | 21.29284 | 0.176811 | 21.4936 | 0.079492 | 2.718282 | 4.52E-16 | 1 |

It can be noticed from (Table 3) that the CSO algorithm is a competitive algorithmfor the modern ones and provides very satisfactory results. In order to perceive the overall performance of the algorithms, they are ranked as shown in (Table 4) according to different benchmark function groups. It can be seen that CSO ranks first in the overall ranking and multimodal test functions. Additionally, it ranks second in unimodal and CEC test functions; (See Figure 4). These results indicate the effectiveness and robustness of the CSO algorithm. That being said, these results need to be confirmed statistically. (Table 5) presents the Wilcoxon matched-pairs signed-rank test for all test functions. In more than 85% of the results, P-value is less than 0.05%, which proves that the results are significant and we can reject the null hypothesis that there's no difference between the means. It is worth mentioning that the performance of CSO can be further evaluated by comparing it against other new algorithms such as Donkey and Smuggler Optimisation Algorithm [111], Modified Grey Wolf Optimiser [112], BSA and its variants [113], WOA and its variants [114], other modified versions of DA [115], etc.

Table 4: Ranking of CSO algorithm compared to the modern Meta-heuristic algorithms

| Test Functions | Ranking CSO | Ranking DA | Ranking BOA | Ranking FDO |
|---|---|---|---|---|
| F1 | 2 | 4 | 3 | 1 |
| F2 | 2 | 4 | 1 | 3 |
| F3 | 2 | 4 | 1 | 3 |
| F4 | 3 | 4 | 1 | 2 |
| F5 | 1 | 4 | 2 | 3 |
| F6 | 3 | 4 | 2 | 1 |
| F7 | 2 | 3 | 1 | 4 |
| F8 | 2 | 3 | 4 | 1 |
| F9 | 2 | 3 | 4 | 1 |
| F10 | 4 | 3 | 2 | 1 |
| F11 | 3 | 4 | 1 | 2 |
| F12 | 2 | 3 | 1 | 4 |
| F13 | 3 | 2 | 1 | 4 |
| F14 | 1 | 2 | 3 | 4 |
| F15 | 2 | 4 | 1 | 3 |
| F16 | 1 | 2 | 4 | 3 |
| F17 | 3 | 4 | 2 | 1 |
| F18 | 3 | 2 | 4 | 1 |
| F19 | 2 | 3 | 4 | 1 |
| F20 | 1 | 2 | 4 | 3 |
| F21 | 1 | 2 | 3 | 4 |
| F22 | 1 | 2 | 4 | 3 |
| F23 | 1 | 2 | 3 | 4 |
| Cec01 | 3 | 4 | 2 | 1 |
| Cec02 | 3 | 4 | 2 | 1 |
| Cec03 | 2 | 3 | 4 | 1 |
| Cec04 | 2 | 3 | 4 | 1 |
| Cec05 | 3 | 2 | 4 | 1 |
| Cec06 | 2 | 1 | 3 | 4 |
| Cec07 | 2 | 3 | 4 | 1 |
| Cec08 | 1 | 2 | 4 | 3 |
| Cec09 | 2 | 3 | 4 | 1 |
| Cec10 | 3 | 2 | 4 | 1 |
| TOTAL | 70 | 97 | 91 | 72 |
| OVERALL RANKING | **2.121212** | 2.939394 | 2.757576 | 2.181818 |
| F1-F7 SUBTOTAL | 15 | 27 | **11** | 17 |
| F1-F7 RANKING | 2.142857 | 3.857143 | **1.571429** | 2.428571 |
| F8-F23 SUBTOTAL | **32** | 43 | 45 | 40 |
| F8-F23 RANKING | **2** | 2.6875 | 2.8125 | 2.5 |
| CEC01-CEC10 SUBTOTAL | 23 | 27 | 35 | **15** |
| CEC01-CEC10 RANKING | 2.3 | 2.7 | 3.5 | **1.5** |

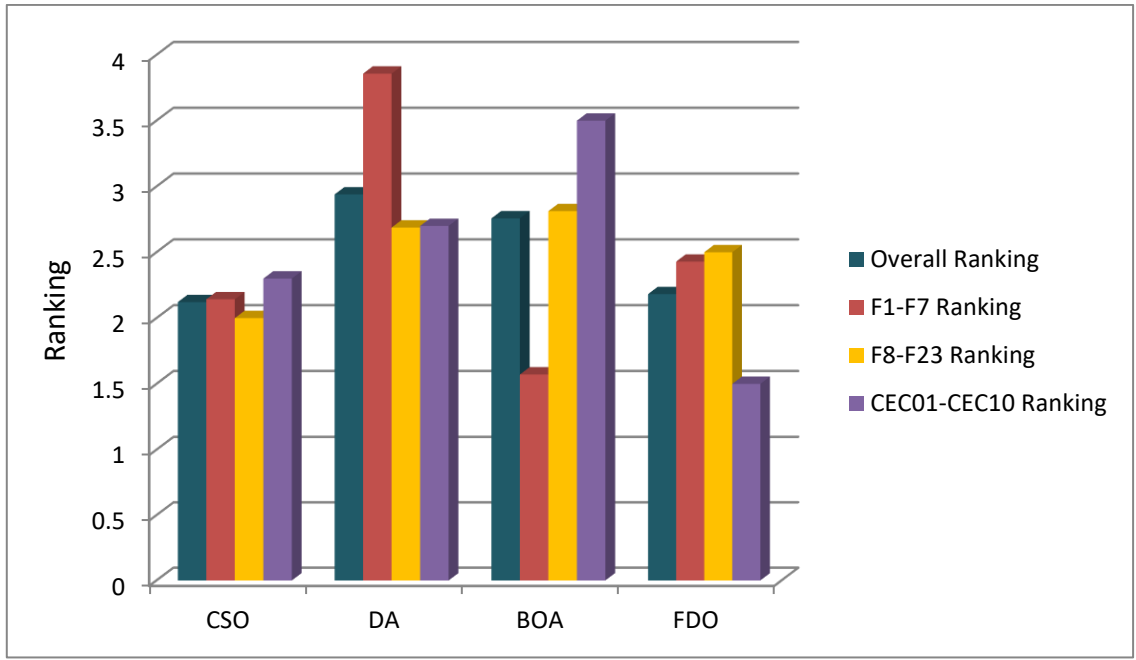

Figure 4: Ranking of algorithms according to different groups of test functions.

Table 5: Wilcoxon matched-pairs signed-rank test

| test functions | CSO vs. DA | CSO vs. BOA | CSO vs. FDO |
|---|---|---|---|
| F1 | <0.0001 | <0.0001 | <0.0001 |
| F2 | <0.0001 | <0.0001 | 0.0003 |
| F3 | <0.0001 | <0.0001 | 0.2286 |
| F4 | <0.0001 | <0.0001 | <0.0001 |
| F5 | <0.0001 | 0.0879 | 0.0732 |
| F6 | 0.0008 | 0.271 | <0.0001 |
| F7 | 0.077 | <0.0001 | <0.0001 |
| F8 | 0.586 | N/A | <0.0001 |
| F9 | 0.2312 | 0.3818 | <0.0001 |
| F10 | 0.0105 | <0.0001 | <0.0001 |
| F11 | <0.0001 | <0.0001 | 0.0002 |
| F12 | 0.4 | <0.0001 | <0.0001 |
| F13 | <0.0001 | <0.0001 | 0.0185 |
| F14 | 0.4 | <0.0001 | 0.0003 |
| F15 | 0.0032 | 0.0004 | 0.9515 |
| F16 | <0.0001 | N/A | <0.0001 |
| F17 | <0.0001 | <0.0001 | <0.0001 |
| F18 | <0.0001 | <0.0001 | <0.0001 |

| F19 | 0.2109 | N/A | 0.6554 |
|-----|--------|-----|--------|
| F20 | 0.0065 | N/A | <0.0001 |
| F21 | 0.0057 | <0.0001 | <0.0001 |
| F22 | 0.1716 | <0.0001 | <0.0001 |
| F23 | <0.0001 | <0.0001 | <0.0001 |
| cec01 | <0.0001 | <0.0001 | <0.0001 |
| cec02 | 0.001 | <0.0001 | <0.0001 |
| cec03 | 0.0102 | <0.0001 | <0.0001 |
| cec04 | 0.0034 | <0.0001 | <0.0001 |
| cec05 | 0.1106 | <0.0001 | <0.0001 |
| cec06 | 0.0039 | 0.0007 | <0.0001 |
| cec07 | 0.0002 | <0.0001 | <0.0001 |
| cec08 | 0.0083 | <0.0001 | <0.0001 |
| cec09 | 0.115 | <0.0001 | <0.0001 |
| cec10 | 0.0475 | <0.0001 | <0.0001 |

## 7. Conclusion and future directions

Cat Swarm Optimization (CSO) is a metaheuristic optimization algorithm proposed originally by Chu et al. in 2006. Henceforward, many modified versions and applications of it have been introduced. However, the literature lacks a detailed survey in this regard. Therefore, this paper firstly addressed this gap and presented a comprehensive review including its developments and applications.

CSO showed its ability in tackling different and complex problems in various areas. However, just like any other meta-heuristic algorithm; the CSO algorithm possesses strengths and weaknesses. The Tracing mode resembles the global search process while the seeking mode resembles the local search process. This algorithm enjoys a significant property for which these two modes are separated and independent. This enables researchers to easily modify or improve these modes and hence achieve a proper balance between exploration and exploitation phases. In addition, fast convergence is another strong point of this algorithm, which makes it a sensible choice for those applications that require quick responses. However, the algorithm has a high chance of falling into local optima, known as premature convergence, which can be considered as the main drawback of the algorithm.

Another concern was the fact that the CSO algorithm was not given a chance to be compared against new algorithms since it has been mostly measured up against PSO and GA algorithms in the literature. To address this, a performance evaluation was conducted to compare CSO against three new and robust algorithms. For this, 23 traditional benchmark functions and 10 modern benchmark functions were used. The results showed the outperformance of the CSO algorithm, in which it ranked first in general. The significance of these results was also confirmed by statistical methods. This indicates that CSO is still a competitive algorithm in the field.

In the future, the algorithm can be improved in many aspects; for example, different techniques can be adapted to the tracing mode in order to solve the premature convergence problem, or Changing the MR parameter into a dynamic parameter to properly balance between exploration and exploitation phases.

## Conflicts of Interest

The authors declare that there is no conflict of interest regarding the publication of this paper.

## Acknowledgments

This is research did not receive any specific grant from funding agencies in the public, commercial, or not-for-profit sectors